# A Fuzzy Reinforcement LSTM-based Long-term Prediction Model for Fault Conditions in Nuclear Power Plants

Siwei Li[a,b], Jiayan Fang[a], Yichun Wu[a,*], Wei Wang[b], Chengxin Li[c], Jiangwen Chen[b]

[a]*College of Energy, Xiamen University, Xiamen 361102, China*
[b]*State Nuclear Power Automation System Engineering Company, Shanghai 200233, China*
[c] *Tongji University, Shanghai 201804, China*

## Abstract

Early fault detection and timely maintenance scheduling can significantly mitigate operational risks in NPPs and enhance the reliability of operator decision-making. Therefore, it is necessary to develop an efficient Prognostics and Health Management (PHM) multi-step prediction model for predicting of system health status and prompt execution of maintenance operations. In this study, we propose a novel predictive model that integrates reinforcement learning with Long Short-Term Memory (LSTM) neural networks and the Expert Fuzzy Evaluation Method. The model is validated using parameter data for 20 different breach sizes in the Main Steam Line Break (MSLB) accident condition of the CPR1000 pressurized water reactor simulation model and it demonstrates a remarkable capability in accurately forecasting NPP parameter changes up to 128 steps ahead (with a time interval of 10 seconds per step, i.e., 1280 seconds), thereby satisfying the temporal advance requirement for fault prognostics in NPPs. Furthermore, this method provides an effective reference solution for PHM applications such as anomaly detection and remaining useful life prediction.

## 1 Introduction

The objective of Prognostics and Health Management (PHM) is to continuously

monitor the operational condition of critical systems and identify any deviations from normal functioning. Effective PHM methods can mitigate the occurrence of severe accidents and enhance system safety (Shi et al., 2024). It harnesses historical, present, and future knowledge, information, and data to estimate the health of components and predict their status, enabling fault identification and prediction of fault patterns (Nguyen et al., 2020).

Early fault detection and timely maintenance scheduling can significantly mitigate operational risks and enhance the reliability of operator decision-making in nuclear power plants (NPPs). During plant operation, human operators are responsible for monitoring a multitude of instrument signals and applying them to corresponding procedures based on the plant's status (Kim et al., 2020). However, in the event of emergencies or anomalies, operators are faced with extremely limited response time. Given the intricate nature of accidents and the pressing urgency, there is a propensity for operators to make erroneous decisions (Wang et al., 2016). Therefore, it is imperative to develop efficient PHM multi-step long-term prediction models that can facilitate prompt maintenance operations.

To achieve long-term parameter prediction, employed methodologies encompass knowledge-based approaches, statistical model-based methods, and data-driven prediction techniques. Knowledge-based approaches rely on extensive expert knowledge derived from physics equations, material properties, reliability theory, etc., to accurately model diverse systems.

For example, Godino et al. (2023) employed a multi-domain computational approach to model the RD-14M steam generator, utilizing two-fluid Eulerian methods for simulating water-steam flow and investigating both steady-state and transient events. El-Morshedy et al. (2023) established a mathematical model to analyze the thermal performance of the VVER-1200 pressurized water reactor (PWR) under normal operating conditions. Through analytical solutions of energy and heat conduction equations analytically, the authors predicted temperature trends in coolant, cladding, and fuel. However, knowledge-based methods are impractical for real-time prediction due to their time-consuming "compute-assess-correct" cycles. Moreover, the nonlinear and multidimensional nature of NPPs renders knowledge-based approaches unsuitable for managing plant data (Kim et al., 2021). This is primarily attributed to the challenge of deriving precise mathematical formulas for nonlinear, multidimensional, time-varying systems.

Statistical-based methods infer system rules by summarizing patterns in historical data, requiring less expert knowledge for system modeling. Commonly employed techniques include Autoregressive Integrated Moving Average (ARIMA) models, logistic regression, etc. For example, Contreras et al. (2003) applied a simple ARIMA model to forecast electricity prices for the following day. (Rocco S, 2013) utilized Singular Spectrum Analysis (SSA) to decompose failure behavior into trend, oscillatory behavior, and noise independent components for multi-step prediction. Hussain et al. (2023) combined extreme value theory with ARIMA models to predict collision risks at signalized intersections over the next 30-35 minutes. Jiang et al. (2024) proposed three prediction methodologies, namely auto-MPSIC, IV-MPSIC, and MSEI-MPSIC, based on Multi-Period Sequence Indicator Combination (MPSIC) technique for short-term prediction with limited sample data. However, statistical-based methodologies are constrained by stringent assumptions and rely on parameterized models, thereby limiting their applicability to independent datasets. Their effectiveness diminishes when confronted with complex nonlinear problems, as they tend to introduce significant uncertainties (Zeng et al., 2024).

With the widespread integration of monitoring systems in complex industrial processes, data-driven methods have gained significant traction, enabling swift capture of vast amounts of monitoring data that promptly reflects the operational status and fault information of the system, thereby facilitating fault prediction (Xiao et al., 2023). Data-driven methodologies can be categorized into classical machine learning methods and deep learning approaches. Representative classical machine learning models encompass decision trees (DT), support vector machines (SVM), and random forests (RF). Recently, highly complex neural network architectures and deep structures, as a significant branch of machine learning, have been extensively employed in diverse time series prediction domains when combined with traditional methods. For instance, Zeng et al. (2024) proposed the STL-transformer-ARIMA model, which integrates statistical learning and deep learning, to enable short-term prediction of aviation fault events. Chehade et al. (2024) introduced SeqOAE, a sequence-to-sequence deep learning model, to achieve robust long-term time series forecasting in variable population sizes.

In the field of nuclear power, Santosh et al. (2009) proposed an artificial neural network-based method for rapid identification of transients and initiation of corrective measures during accident scenarios in the 220 MWe Indian PHWR. Moshkbar-Bakhshayesh (2019) employed a range of supervised learning methods to predict transients for critical parameters in NPPs, including reactor inlet flow rate and steam generator pressure. She et al. (2021) employed a combination of Convolutional Neural Network (CNN), Long Short-Term Memory (LSTM) neural network, and Convolutional LSTM to achieve short-term prediction of coolant loss in NPPs. While short-term fault prediction does indeed facilitate early warning for operators, it falls short in providing operators with adequate emergency response time or a foundation for decision-making. Therefore, it is imperative for NPP PHM prediction models to focus more on long-term (or multi-step) prediction.

The multi-step prediction interval is typically set to match the data sampling interval. For instance, Nguyen et al. (2020) utilized a tree-structured Parzen estimation Bayesian optimization algorithm combined with LSTM, employing a sampling interval of 3 days, to forecast the narrow-range water level of Steam Generators (SGs) 15 steps ahead (equivalent to 45 days). Bae et al. (2021) proposed a long-term prediction model for emergency situations, employing a sampling interval of 30 seconds, thereby facilitating the estimation of future trends for 25 selected parameters up to 20 steps ahead (equivalent to a duration of 10 minutes).

Due to the lack of information, uncertainty, or cumulative errors, achieving multi-step prediction poses a formidable challenge (Song et al., 2020). The aforementioned scholars did not address the issue of error accumulation during the prediction process in their study on multi-step prediction. These accumulated errors serve as significant factors contributing to diminished accuracy of prediction models and erroneous fault trend predictions.

Therefore, this study addresses the optimization of the multi-step prediction problem and the accumulation of errors during the prediction process by formulating it as a reinforcement learning problem within a Multiple-Input Multiple-Output (MIMO) framework (Ben Taieb et al., 2012). After making predictions, the expert fuzzy assesses the model's prediction strategy performance for this round and provides the evaluation score as delayed rewards feedback to the model. The proposed Reinforcement LSTM model based on expert fuzzy evaluation is validated utilizing data from the CPR1000 NPP simulator, which automatically adjust prediction trends. The proposed model is capable of forecasting unit parameter changes up to 128 steps in advance, with a time interval of 10 seconds per step. This meets the temporal requirement for timely fault warning in NPPs.

The remainder of the paper is organized as follows. Theoretical principles underlying the proposed model are outlined in Section 2. The internal structure of the model is described in Section 3. In Section 4, the algorithm model is validated employing data from the CPR1000 NPP simulator. Finally, a conclusion and outlook for the paper are presented.

| Nomenclature | |
|---|---|
| AM | Attention Mechanism |
| ARIMA | Autoregressive Integrated Moving Average |
| BiLSTM | Bi-directional Long Short-Term Memory |
| CNN | Convolutional Neural Networks |
| CVAE | Conditional Variational Auto Encoder |
| DT | Decision Tree |
| DTW | Dynamic Time Warping |
| LOCA | Loss of Coolant Accident |
| LSTM | Long Short Time Memory |
| MC Dropout | Monte-Carlo Dropout |
| MIMO | Multiple Input -Multiple Output |
| MSLB | Main Steam Line Break |
| NPP | Nuclear Power Plant |
| PHM | Prognostics and Health Management |
| RF | Random Forest |
| RNN | Recurrent Neural Network |

| SAM | Similarity Aggregation Method |
| Soft-DTW | Soft Dynamic Time Warping |
| SSA | Singular Spectrum Analysis |
| SVM | Support vector machines |
| TDI | Time Distortion Index |

Table 1 Statistics of parameter prediction methods proposed in recent years

| | Author | Method | Application Scenario | Types of Prediction |
|---|---|---|---|---|
| Knowledge-based methods | Dario M. Godino (2022) | Multi-domain computational modeling | Steady-state nominal events and transient events investigation of steam generators | Transient prediction |
| | Salah El-Din El-Morshedy et al. (2023) | Thermal-hydraulic modeling | VVER-1200 pressurized water reactor | Transient prediction |
| Statistical methods | Javier Contreras (2003) | Simple ARIMA model | Next-day electricity price prediction | Short-term prediction |
| | Claudio M. (2013) | Singular Spectrum Analysis (SSA) | Turbocharger and automotive engine failure behavior prediction | Long-term prediction |
| | Fizza Hussain (2023) | ARIMA model combined with Extreme Value Theory | Signalized intersection collision risk prediction | Short-term prediction |
| | Hongyan Jiang (2023) | MPSIC model | Trend prediction with small-sample data | Short-term prediction |
| Data-driven methods | Hang Zeng (2023) | STL-transformer-ARIMA model | Aviation failure event prediction | Short-term prediction |
| | Abdallah Chehade (2024) | SeqOAE model | Nonlinear time series data prediction with variable population size | Long-term prediction |
| | Santosh et al. (2009) | Artificial Neural Network (ANN) | Rapid identification of transients and initiation of corrective measures during accident scenarios in the 220 MWe Indian PHWR | Short-term prediction |
| | Moshkbar-Bakhshayesh (2019) | Supervised learning methods models (such as FFBP, CFFNN, etc.) | Transient prediction for critical parameters of NPPs (such as reactor inlet flow rate) | Transient prediction |
| | She et al. (2021) | Convolutional LSTM | Short-term prediction of coolant loss in NPPs | Short-term prediction |
| | Nuguyen et al. (2020) | Bayesian optimization algorithm combined with LSTM model | Prediction of narrow-range water level in Steam Generators (SGs) | Short-term prediction |
| | Bae et al. (2021) | Multi-layer Recursive LSTM model | Multi-parameter multi-step prediction under emergency conditions | Multi-step prediction |

# 2 Methods

The present chapter will introduce the principal methodologies employed in the proposed model, encompassing BiLSTM, expert fuzzy evaluation methodology, and assessment indicators such as DTW and TDI. We employ multi-layer BiLSTM to extract sequential features from temporal data and capture their temporal dependencies.

The expert fuzzy evaluation method is utilized to select reinforcement learning indicators for the model and integrate them into a fuzzy evaluation module. Ultimately, MC Dropout is utilized to estimate the confidence interval of the prediction results.

**2.1 Long Short-Term Memory Neural Network (LSTM)**

The LSTM model represents a significant advancement over the conventional RNN model. Traditional RNN models exhibit limited memory capacity, and their predictive capability significantly deteriorates as the input sequence length increases. Early approaches to tackle sequence-related learning problems were either problem-specific or lacked scalability for long-term dependencies. The LSTM model, however, exhibits both generality and effectiveness in capturing long-term dependencies, thereby circumventing the optimization challenges encountered by simple recurrent neural networks (John & Stefan, 2001). The LSTM model proposed by Graves and Schmidhuber (2005), commonly employed in the literature, is known as the vanilla LSTM. The fundamental concept revolves around the regulation of information flow into and out of memory cells via three nonlinear gating units, namely the input gate, forget gate, and output gate (Greff et al., 2017). A conventional LSTM neural network solely captures the dependency of the current state on the previous state, disregarding backward dependencies. Consequently, it fails to fully exploit the hidden information in the data that pertains to backward context (Huang et al., 2019). However, incorporating both past and future context information proves effective for many sequence learning tasks (Khan & Yairi, 2018).

The BiLSTM, which is an extension of the forward LSTM, incorporates a backward LSTM as well. The rationale behind this approach is to input the same sequence into both the forward and backward LSTMs, concatenate their hidden layers, and connect them to the output layer for prediction. Consequently, in this study, we employ the BiLSTM model that effectively captures bidirectional dependency relationships.

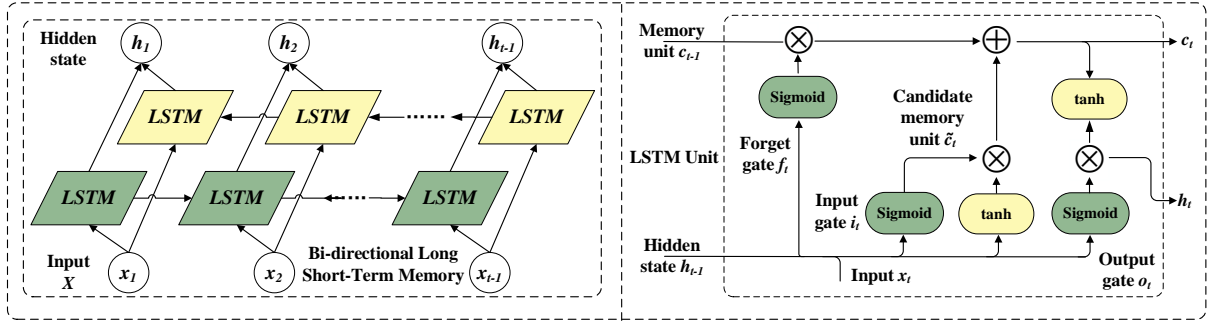

**Fig. 1.** The internal structure of BiLSTM and LSTM units.

The output of the n$^{th}$ BiLSTM is represented as shown in equation (1), where $\overrightarrow{h_t}$ denotes the forward hidden state, and $\overleftarrow{h_t}$ denotes the backward hidden state.

$$h_t^{BiLSTM} = [\overrightarrow{h_t} \oplus \overleftarrow{h_t}] \qquad (1)$$

### 2.2 Soft-DTW and TDI

Dynamic Time Warping (DTW) is a method utilized for quantifying the similarity between two time series by aligning them temporally and calculating their resemblance (Rakthanmanon et al., 2012). However, due to its non-differentiability, DTW cannot be directly employed in neural networks.

Soft-DTW( $DTW_\gamma$ ) addresses the issue of non-differentiability in DTW by facilitating the comparison of similarity or distance between time series of varying lengths (Cuturi & Blondel, 2017). It enables the establishment of one-to-many or many-to-many matches between subsequences within two long sequences, thereby minimizing the overall distance between them.

For two sequences $x = (x_1, x_2, \ldots, x_n) \in R^{p*n}$ and $y = (y_1, y_2, \ldots, y_n) \in R^{p*m}$, let's define their cost matrix $\Delta(x, y) := [\delta(x_i, y_i)]_{ij} \in R^{n*m}$, where $\delta$ represents a differentiable cost function representing the Euclidean distance between corresponding points. Let $R = [r_{i,j}]$, where $R \in R^{n*m}$, be the set of cumulative costs along the path. For DTW, its dynamic programming formulation can be expressed as follows:

$$r_{i,j}^{DTW} = \delta_{i,j} \cdot \omega_{i,j} + \min\{r_{i,j-1}, r_{i-1,j-1}, r_{i-1,j}\} \qquad (3)$$

The "$min$" component in the original formula is substituted with "$min^\gamma$" within

Soft-DTW framework. The specific formula is as follows:

$$min^{\gamma}\{a_1, a_2, \ldots, a_n\} = \begin{cases} min_{i \leq n} a_i & \gamma = 0 \\ -\gamma \log \sum_{i=1}^{n} e^{\frac{-a_i}{\gamma}} & \gamma > 0 \end{cases} \quad (4)$$

The computation formula for Soft-DTW is thus derived as follows:

$$DTW_{\gamma} = min^{\gamma}\{<A, \Delta(x,y)>, A \in A_{n,m}\}$$
$$= -\gamma \log \left( \sum_{A \in A_{n,m}} e^{-<A,\Delta(x,y)>/\gamma} \right) \quad (5)$$

Le Guen and Thome (2019) integrated Soft-DTW with their proposed time loss TDI to formulate the DILATE differentiable loss function, which is employed for investigating time series prediction and multi-step prediction problems of non-stationary signals.

$$TDI(\hat{y}_i, \dot{y}_i) = \langle \arg\min_{A \in A_{k,k}} \langle A, \Delta(\hat{y}_i, \dot{y}_i) \rangle \rangle \quad (6)$$

The Soft-DTW method is utilized to assess the similarity in shape of predicted parameter trends, while TDI is utilized to quantify the extent of temporal deviation in these trends.

**2.3 Expert fuzzy evaluation**

The fuzzy comprehensive evaluation method is a comprehensive assessment approach grounded in the principles of fuzzy mathematics, enabling the conversion of qualitative evaluations into quantitative assessments through the utilization of fuzzy mathematics membership theory. This methodology effectively addresses intricate and indeterminate problems, rendering it suitable for resolving diverse nonlinear issues.

The evaluation of multi-step predictions involves a range of indicators, posing challenges in discerning the relative merits of each indicator for the problem at hand. Therefore, adopting fuzzy linguistic terms instead of precise numerical values to select evaluation indicators offers a more practical approach.

Given the variations in experts' professional backgrounds and knowledge levels,

their evaluations exhibit distinct individual styles. Therefore, selecting an appropriate aggregation method is crucial in fuzzy evaluation approaches. Commonly employed methods include arithmetic mean operator (Detyniecki, 2000), linear opinion pool (Clemen & Winkler, 1999), maximum-minimum Delphi method (Ishikawa et al., 1993), similarity aggregation method (SAM) (Hsi-Mei & Chen-Tung, 1996), and fuzzy analytic hierarchy process (FAHP) (Yazdi & Kabir, 2017). Among these methods, only SAM comprehensively considers expert weights and consensus.

Therefore, this study adopts for an improved SAM approach to consolidate expert fuzzy opinions, with the aim of attaining more robust and dependable outcomes. In practical terms, experts are drawn from diverse professional backgrounds, educational qualifications, and experiences. Higher professional positions bolster the credibility of their viewpoints by reflecting their extensive theoretical knowledge and substantial practical expertise. Accumulated experience in the field contributes to heightened judgment accuracy. Additionally, educational backgrounds generally indicate the depth of expertise accumulated in a specific domain (Bu et al., 2023). This paper comprehensively considers these three factors when determining expert weights.

The specific steps for improving SAM are as follows (Guo et al., 2021):

(1) Convert linguistic terms into corresponding fuzzy numbers. Experts utilize the predetermined fuzzy terms presented in Table 3 to express their opinions. Based on the conversion relationships illustrated in Table 3, the fuzzy terms are transformed into respective fuzzy numbers.

(2) Calculate the consistency of each pair of expert opinions $S_{uv}(\tilde{R}_u, \tilde{R}_v)$, where $S_{uv}(\tilde{R}_u, \tilde{R}_v)$ is in the range [0,1], and is defined as follows:

$$S_{uv}(\tilde{A}, \tilde{B}) = 1 - \frac{1}{4}\sum_{i=1}^{4}|a_i - b_i| \qquad (7)$$

The higher the value, the greater the consistency of opinions between the two experts.

(3) Calculate the degree of weighted (absolute) agreement degree (WA).

Table 2 Weighted Criteria for Expert Ratings

| | Category | weight |
|---|---|---|
| Professional position | Professor | 4 |
| | Associate Professor/Senior Engineer | 3 |
| | Assistant Professor/Engineer | 2 |
| | Technician/Operator | 1 |
| Work experience | ＞30 years | 4 |
| | 20-29 | 3 |
| | 10-29 | 2 |
| | ＜10 years | 1 |
| Education level | PhD | 2 |
| | Master | 1.5 |
| | Bachelor | 1 |

Table 3 Fuzzy number sets of the scale

| Fuzzy Evaluation | | Fuzzy Set |
|---|---|---|
| Very Low | VL | (0.00,0.00,0.10,0.20) |
| Low | L | （0.10,0.25,0.25,0.40） |
| Moderate | M | （0.30,0.45,0.55,0.70） |
| High | H | （0.60,0.75,0.75,0.90） |
| Very High | VH | （0.80,0.90,0.90,1.00） |

## 2.4 MIMO

The term "multi-step ahead prediction" refers to predicting of $\{x_{t+1}, x_{t+2} ... x_{t+H}\}$ given the observed values $\{x_1, x_2 ... x_t\}$ and a forecast horizon of H time steps. In order to achieve multi-step ahead prediction, recursive prediction strategies, direct prediction strategies, and MIMO (Multiple Input-Multiple Output) strategies are commonly employed (Ben Taieb et al., 2010).

To achieve long-term time series prediction using recursive methods, it is necessary to iteratively perform one-step ahead predictions. For predicting H steps ahead, this iteration needs to be repeated H times. Common examples of iterative methods include recurrent neural networks (RNNs) (Williams & Zipser, 1989) and local learning iterative techniques (Farmer & Sidorowich, 1987).On the contrary, the direct method involves estimating H one-step ahead prediction models in order to accomplish H-step ahead forecasting. When there is a significant time gap between two consecutive steps, it becomes necessary to make modeling adjustments based on their interdependency (Sorjamaa et al., 2007). The first two methods can be regarded as single-output strategies since they model the data as functions with multiple inputs but a single output. The Multiple Input-Multiple Output (MIMO) strategy, on the other hand, adheres to the fundamental principle of maintaining stochastic dependencies among predicted values, thereby preserving the characteristic representation of time series data. This approach avoids making assumptions about conditional independence and issues related to error accumulation encountered in recursive methods.

When conducting predictions, the MIMO strategy enables simultaneous prediction of H steps.

$$\{\hat{x}_{t+1}, \hat{x}_{t+2}, \ldots\ldots \hat{x}_{t+H}\} = f_1\{x_{t-d+1}, x_{t-d+2}, \ldots\ldots, x_t\} \tag{8}$$

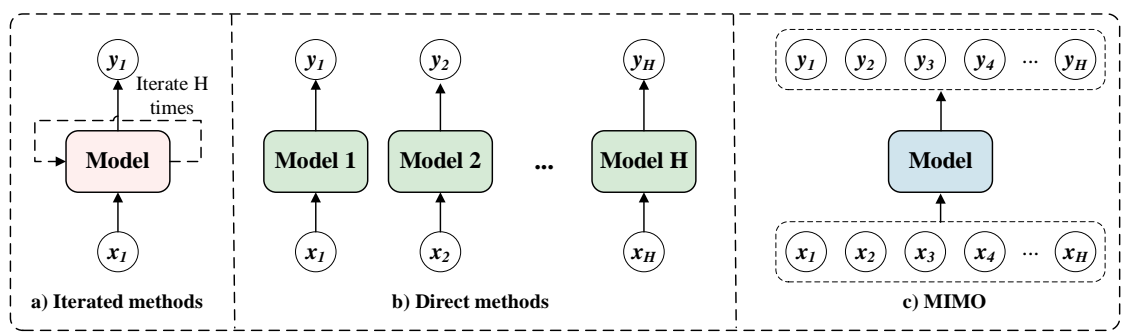

**Fig. 2.** Comparative Illustration of Prediction Methods

## 2.5 Sliding Window

The horizontal axis in Fig. 2 represents the feature dimension, while the vertical axis represents the time dimension. Different features are distinguished by varying colors. A sliding window of length $L$ is used to encapsulate the sensor signal sequence. At each time step, a new sliding window is generated until reaching the end of the time series. $X_i = \{x_1, x_2, x_3, \ldots, x_l, \ldots x_{N-L+1}\}$ denotes a sequence of sliding windows within a dataset, where $x_i \in R^{L \times n}$, $n$ signifies the number of features. Herein, $L$ denotes the length of the sliding window, and $N$ represents the total length of time.

A larger sliding window facilitates the inclusion of a greater amount of information, thereby enhancing the capability to extract latent features embedded within sensor signal sequences (Liu et al., 2022).

However, this approach may potentially impede the training speed of the model and introduce heightened computational complexity. Furthermore, as the window length surpasses a certain threshold, there is a possibility that it could adversely impact the accuracy of the model (Huang et al., 2019). In accordance with prior research (Chen et al., 2021), a sliding window length of 40 has been adopted.

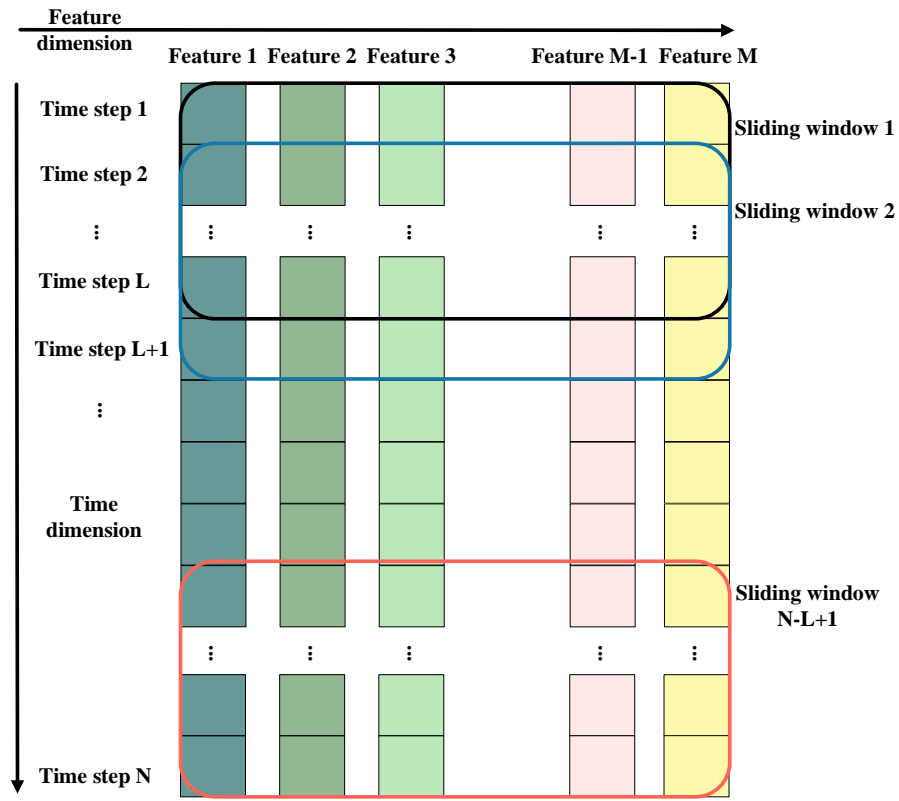

**Fig. 3.** Diagram of the Sliding Window Mechanism

# 3 Proposed long-term prediction algorithm framework

The algorithm proposed in this paper is introduced in this section, as illustrated in Fig. 3. It includes four parts: data preprocessing, multi-step prediction, confidence interval estimation, and post-processing. The overall process can be summarized as follows:

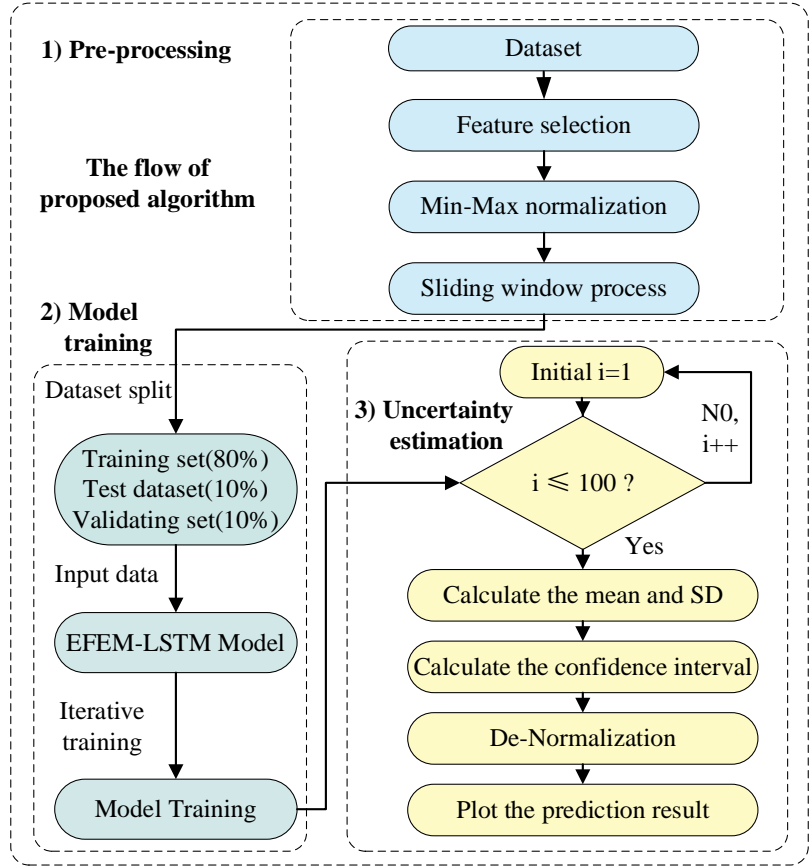

**Fig. 4.** Algorithm Flowchart

## 3.1 Pre-preprocessing function

The purpose of data preprocessing is to transform raw parameters into inputs suitable for neural networks. In practical applications, there may be variations in the scales of numerous raw data, operational parameters, loads, and fault time values. The normalization of the data is therefore necessary prior to training (Zheng et al., 2017).

The raw data in this study was standardized using Z-Score normalization, employing the following formula:

$$x_{input} = \frac{x - \mu}{\sigma} \tag{9}$$

$$\mu = \frac{1}{n}\sum_{i=1}^{n} x_i \tag{10}$$

$$\sigma = \frac{1}{n}\sqrt{\sum_{i=1}^{n}(x_i - \mu)^2} \tag{11}$$

The formula above utilizes $x$ to represent the initial value of the current NPP parameter, while $\mu$ and $\sigma$ denote the calculated mean and standard deviation correspondingly. On the other hand, $x_{input}$ represents the standardized data value, which conforms to a standard normal distribution with a mean of 0 and a standard deviation of 1. Employing normalization for data processing demonstrates remarkable robustness as it does not necessitate data stability and accommodates extreme maximum or minimum values.

## 3.2 Prediction model framework overview

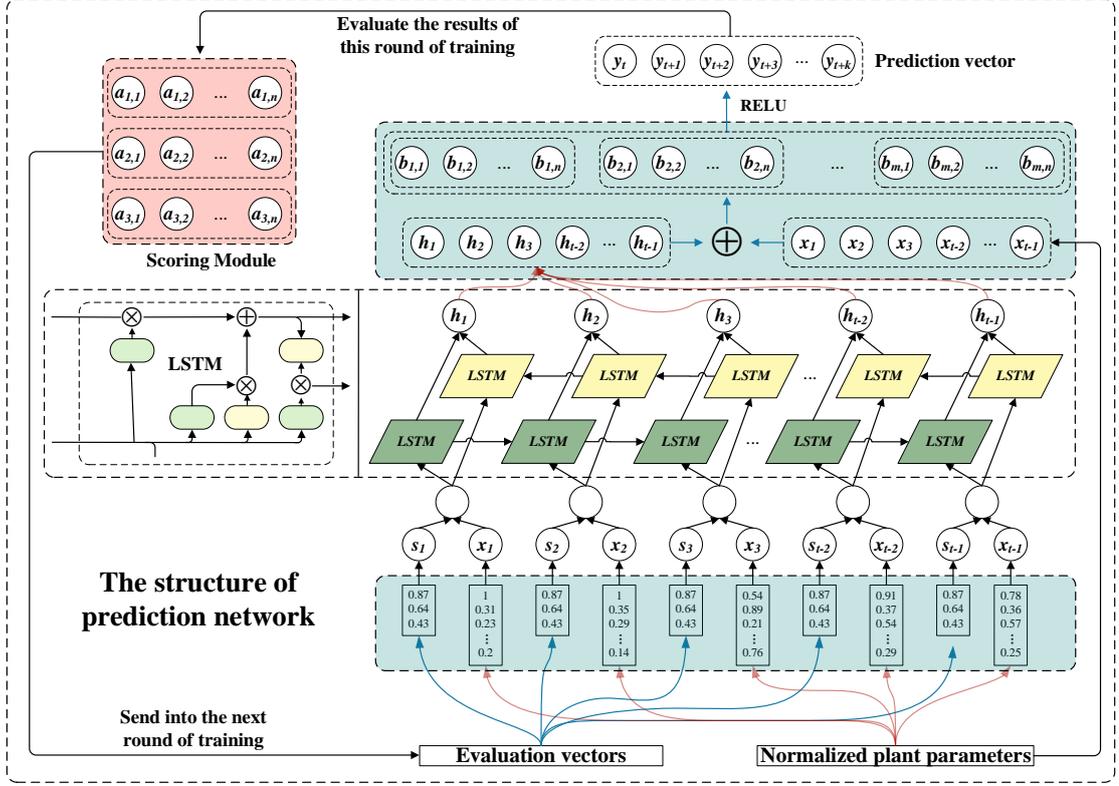

**Fig. 5.** Long-term prediction model framework overview

The multi-step prediction function is based on the MIMO framework, utilizing normalized unit parameters from the preprocessing function and scores obtained from the fuzzy evaluation module as inputs. This function enables accurate forecasting of future trends for the next 128 steps (equivalent to 1280 seconds).

The framework of the multi-step prediction function is illustrated in Fig. 5. For the input multivariate time series data, denoted as $X = [x_1, x_2, ......x_t] \oplus [S_1, S_2, ......, S_t] \in R^{T \times D}$ (where $T$ represents the length of the input time steps and $D$ denotes the number of input parameter features).

The BiLSTM employs a forget gate to selectively discard the contents of the previous cell state $C_{t-1}$, thereby determining which information to retain. Subsequently, an update gate is utilized to selectively incorporate new information into the current cell state. Finally, based on the updated cell state $C_t$, it selectively outputs its current hidden representation $h_t^{BiLSTM}$. ($h_t^{BiLSTM} = \overrightarrow{h_t} \oplus \overleftarrow{h_t}$).

After obtaining the forward and backward hidden information $h_t^{BiLSTM}$, they are jointly fed into the residual neural network along with $X$. The residual neural network estimates the future trend of parameter changes $\hat{Y}$, based on the hidden information $h_t^{BiLSTM}$ and the latent content in $X$. $\hat{Y} = [\hat{y}_{t+1}, \hat{y}_{t+2}, \ldots\ldots, \hat{y}_{t+T}] \in R^T$ (where $T$ is the length of the predicted time steps).

### 3.3 Uncertainty estimation function

The complexity and scale of the NPP system are widely recognized, and even minor modifications to the established system can potentially introduce new defects and result in losses(Yiru et al., 2022). During prediction, neural networks may yield results with significant uncertainty, particularly in intrusion detection systems for critical infrastructure such as NPPs (Linda et al., 2009). Therefore, it is necessary to add the uncertainty information of the prediction results of the model .Uncertainty also plays a pivotal role in the domain of reinforcement learning (Szepesvari, 2010), as it aids the model in evaluating the accuracy of the current strategy and determining optimal methods for exploring the environment. Bayesian models offer a mathematical framework for inferring model uncertainty; however, their computational costs often pose significant challenges. By employing dropout training in deep neural networks as an approximation to Bayesian inference in deep Gaussian processes, we can obtain reliable estimates of model uncertainty without compromising complexity and accuracy (Gal & Ghahramani, 2016). After the training process, we utilize the Monte Carlo (MC) Dropout method to acquire uncertainty information for multi-step prediction.

In this study, we iteratively perform the forward pass 100 times while applying random dropout to the neurons in each layer. Subsequently, by repeating this process for 100 iterations, we derive the mean prediction and confidence interval based on the obtained results.

The protocol for implementing the MC Dropout methodology is outlined as follows:

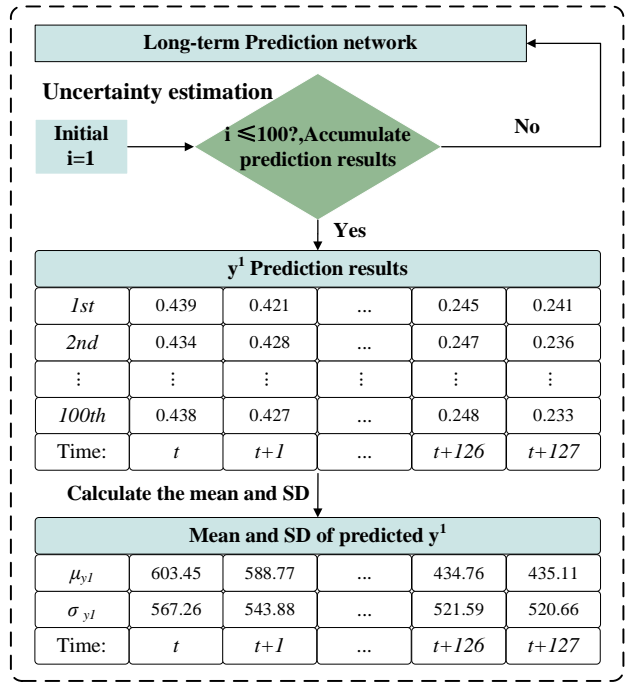

**Fig. 6.** Flowchart of the MC Dropout Method

### 3.4 Post-processing function

The final function performs inverse standardization on the prediction results and maps them onto the image. The formula for inverse normalization is presented as follows:

$$\hat{Y} = Y_\mu + Y * Y_\sigma \tag{12}$$

In the above formula, $\hat{Y}$ and $Y$ represent the inverse standardized prediction values and the original prediction values, respectively. $Y_\mu$ and $Y_\sigma$ denote the computed mean and standard deviation, correspondingly. Finally, the function uses the inverse standardized mean to plot and fills the region between the upper and lower boundaries of the confidence interval.

The validity of the proposed prediction method is demonstrated in this section through simulation data of the MSLB accident condition of the CPR1000 PWR. Given that over 70% of the world's 437 operational reactors by the end of 2022 are PWRs (WNA, 2023), it is crucial to develop a simulation model specifically for a PWR, establish a fault prediction algorithm model, and evaluate its performance.

# 4 Experiments and results

The effectiveness of the proposed prediction method is validated in this section using simulated data from Main Steam Line Break (MSLB) accidents in CPR1000 PWR.

### 4.1 Dataset description

In the event of a Main Steam Line Break (MSLB) accident, an abrupt loss of steam occurs from the secondary loop, resulting in increased heat dissipation from the primary loop system and subsequently reducing the average temperature and pressure of the coolant. In cases where there is a negative moderator temperature coefficient, this decrease in temperature triggers an automatic increase in core reactivity and power to maintain thermal equilibrium between the primary and secondary loop systems, ultimately leading to an emergency shutdown. If the moderator temperature coefficient exhibits a significantly large negative value, there exists a risk of the core returning to criticality (Li, 1997).

The process of the accident can be roughly described in the following two stages: In the first stage, a rupture occurs in the main steam line, resulting in a significant loss of steam. This leads to a rapid increase in both steam flow rate and reactor power as compensation for the false increase in load on the secondary loop. Simultaneously, due to the decrease in average coolant temperature within the primary loop, there is also a corresponding decrease in pressure and water level within the pressurizer. Consequently, this triggers an emergency reactor trip due to over-power protection or pressurizer low-pressure protection, along with turbine trip caused by turbine trip-out of synchronization.

The second stage occurs after the reactor shutdown and turbine trip, but before the isolation of the main steam line. During this period, steam continues to be lost from the rupture, leading to a gradual decrease in average coolant temperature within the primary loop. If the reactor possesses intrinsic characteristics with a negative temperature coefficient, this decline in coolant temperature introduces positive reactivity into the core continuously, thereby reducing the shutdown depth over time. In such circumstances, if the most reactive control rod assembly gets stuck at the top of the core, it is possible for a previously shut-down reactor to return to criticality and reach a certain power level. Consequently, the flux distribution within the core will be significantly distorted, and the cladding of fuel rods located at regions experiencing local power peaks may suffer burnout due to overheating.

In summary, the MSLB accident directly impacts the normal cooling of the reactor, thereby posing significant safety risks to the reactor, potentially leading to r the release of radioactive materials, which poses a grave threat to public safety and environmental integrity. Additionally, it may also cause damage to power plant equipment, thus undermining its economic viability.

Through a comparative analysis of mass and energy release during large break loss of coolant accidents (LBLOCA) and Main Steam Line Break (MSLB) accidents, it is believed that MSLB accidents present the most severe conditions in terms of mass and energy release (Guijun & Yu, 2003). Therefore, it is necessary to forecast the long-term trends of parameters under MSLB accidents. To ensure algorithm effectiveness, training with real data from NPP accident conditions becomes essential. However, such real data is scarce and often classified. Therefore, we employ CPR1000 simulator data for training purposes. The accident scenario of MSLB is simulated, with the rupture size of the main steam line increasing incrementally from 0.005 $m^2$ to 0.13 $m^2$, totaling 20 rupture sizes at intervals of 0.005 $m^2$. Throughout all simulations, the reactor is initially operating at full power (100%FP), and the main steam line break fault injection begins after a delay of 400 seconds.

During the simulation process, data is sampled at intervals of 10 seconds, capturing 78 parameters from the simulation model with each sampling instance. A total of 20,000 samples are collected and subsequently partitioned into training, validation, and test sets in an 8:1:1 ratio.

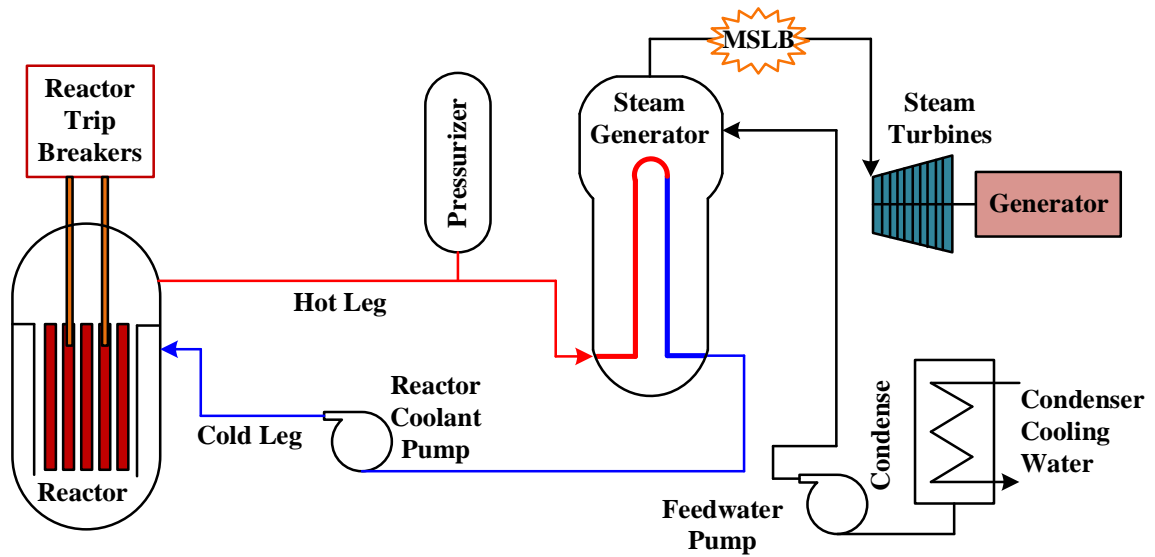

Fig. 7. A block diagram of a representative PWR plant

### 4.2 Implementation

The computing platform specifications are as follows: 6× Xeon E5-2678 v3 CPUs, NVIDIA GeForce RTX 2080 GPU, and 32GB RAM. This work utilizes Python 3.7 as the programming language and employs the PyTorch framework along with multiple Python libraries for algorithm modeling.

## 4.3 Output Parameter Determination and Data Segregation

4.3.1 Output Parameter Determination

Before constructing the model, it is crucial to precisely define the input and output parameters of the model. In the experiment, we identified 24 key parameters as prediction variables based on the International Atomic Energy Agency safety standard . These selected parameters play a crucial role in monitoring operator implementation of emergency operating procedures during MSLB accidents at NPPs.

**Table 4** Prediction Parameters

| Reactor Thermal Power | Pressure of Containment | Temperature of Containment |
|---|---|---|
| Hot-leg #1 Temperature | Hot-leg #2 Temperature | Hot-leg #3 Temperature |
| Cold-leg #1 Temperature | Cold-leg #2 Temperature | Cold-leg #3 Temperature |
| Pressurizer Temperature | Pressurizer Level | Pressurizer Pressure |
| Loop#1 Flow | Loop#2 Flow | Loop#3 Flow |
| SG#1 Pressure | SG#2 Pressure | SG#3 Pressure |
| SG#1 Steam Flow | SG#2 Steam Flow | SG#3 Steam Flow |
| SG#1 Narrow Range Level | SG#2 Narrow Range Level | SG#3 Narrow Range Level |

4.3.2 Optimization of input variables

The objective of feature selection is to extract the most valuable features from a vast pool of collected features based on specific criteria and eliminate redundant ones. Typically, Pearson or Spearman correlation coefficients are commonly employed to establish correlations among variables. The Pearson correlation coefficient reflects the degree of linear association between two continuous variables, while the Spearman coefficient is a non-parametric measure of rank correlation that describes the monotonic relationship between variables (Fernández-Gavilanes et al., 2018). Given that the rate of increase or decrease between two variables is generally not constant and may not follow a linear pattern, we opted for utilizing the Spearman coefficient in order to capture their monotonic relationship.

The formula is as stated below：

$$\rho_s = \frac{\sum(R_i - \bar{R})\sum(S_i - \bar{S})}{\sqrt{\sum(R_i - \bar{x})^2 \sum(S_i - \bar{S})^2}} \quad (13)$$

When $\rho_s = 1$, a strong positive linear correlation is observed, indicating that an increase in one variable corresponds to a proportional increase in the other variable. When $\rho_s > 0$ but not 1, there still exists a positive correlation between the variables; however, the increase in one variable does not consistently result in an equal increase in the other variable. During feature selection, we consider $|\rho_s| \geq 0.4$ as the screening condition to identify features with significant correlations. The model is trained using features that exhibit moderate strength correlations as mentioned above.

4.3.3 Hyperparameters optimization

In order to minimize the impact of hyperparameter variations on model result comparisons, consistent hyperparameters were employed for all network models in this study. The specific model hyperparameters are presented in Table 5, and these values were determined through iterative experimentation.

**Table 5** Model hyperparameters

| Initial learning rate | Weight decay | Batch | Epochs | Optimizer |
|---|---|---|---|---|
| 0.001 | 1e-4 | 64 | 1000 | Adam |
| Hidden states | Layers | Dropout | Input time steps | Output time steps |
| 128 | 2 | 0.2 | 40 | 128 |

**4.4 Expert fuzzy evaluation and defuzzification**

The quantitative assessment of reinforcement learning metrics on prediction performance presents a significant challenge in general, as it is often difficult to anticipate their ultimate impact on the model's predictions.

Therefore, in order to enhance the reliability of predictions and enable adaptive adjustment of the model's prediction trend during training, we employ the fuzzy comprehensive evaluation method for qualitative assessment of learning indicators, which are subsequently converted into quantitative evaluations using the improved SAM algorithm. The specific process is as follows: (1) Ten common indicators are selected to evaluate the predictive performance of the model in time series forecasting, encompassing measures such as dynamic time warping distance and Euclidean distance. (2) Five experts specializing in reliability assessment are invited to qualitatively assess these indicators, which are then transformed into quantitative evaluations using the improved SAM method. (3) The temporal, spatial, and trend similarities between two time series are assessed as three distinct indices. (4) The evaluation score from the previous round is incorporated into iterative training of the model. Expert information and the fuzzy evaluation results for the 10 indicators are presented in Tables 6 and 7, respectively, while Fig. 7 illustrates the mapping of expert fuzzy evaluation outcomes.

Table 6 Weighting criteria and score of experts

| Expert | Professional position | Experience | Education level | Score | Weight |
| --- | --- | --- | --- | --- | --- |
| Expert 1 | Professor | 22 | Ph.D. | 9 | 0.286 |
| Expert 2 | Associate professor | 16 | Ph.D. | 7 | 0.222 |
| Expert 3 | Assistant Professor | 7 | Ph.D. | 5 | 0.159 |
| Expert 4 | Senior Engineer | 12 | Undergraduate | 6 | 0.190 |
| Expert 5 | Engineer | 3 | Master | 4.5 | 0.143 |

Table 7 The outcomes of expert evaluation using fuzzy language analysis

| Metrics | Expert 1 | Expert 2 | Expert 3 | Expert 4 | Expert 5 | Score |
| --- | --- | --- | --- | --- | --- | --- |
| MAE (Mean Absolute Error) | M | M | M | H | H | 0.588 |
| MAPE (Mean Absolute Percentage Error) | M | M | M | H | H | 0.588 |
| MSE (Mean Square Error) | H | M | M | M | H | 0.601 |
| RMSE (Root Mean Square Error) | H | M | H | M | VH | 0.675 |
| SSE (The sum of squares due to error) | M | L | M | L | M | 0.400 |
| Edit Distance | M | M | L | M | M | 0.459 |
| DTW (Dynamic Time Warping) | VH | H | M | H | VH | 0.780 |
| TDI (Temporal Distortion Index) | M | H | H | M | H | 0.641 |
| Cross-Correlation Function | M | H | M | M | M | 0.548 |
| LCS (Longest Common Subsequence) | H | M | M | M | H | 0.601 |

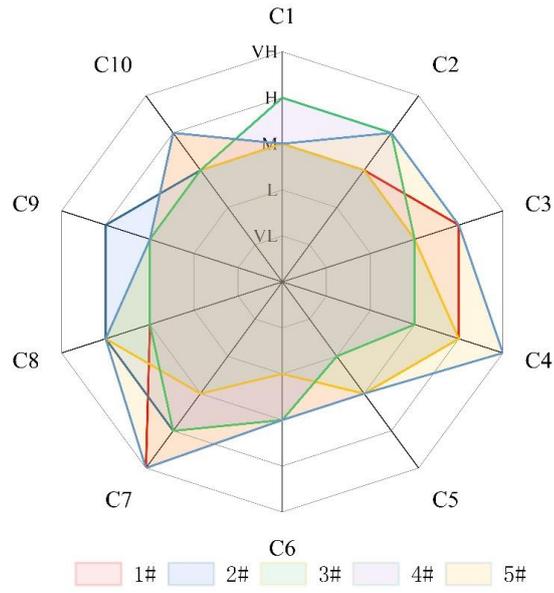

**Fig. 8.** The outcomes of expert evaluation using fuzzy language analysis

**4.5 Validation**

4.5.1 Model prediction results

During the experiment, due to the lack of variation trends for each parameter under MSLB accidents in NPPs from data obtained during stable operating conditions and considering a break inserted 400 s after simulation start, only stable operating condition data from 400 s prior to break insertion were selected as input. Historical data preceding this period was excluded, resulting in a sliding window length of 40.

The proposed EFEM-BiLSTM method is employed for the training and validation of each parameter. Specifically, a total of 16 accident scenarios with 16000 sets of historical data are utilized to train the model in forecasting the subsequent change trend of each parameter within 1280 s following the occurrence of an MSLB accident. In all scenarios, the fault is introduced 400 s after the simulation commences; therefore, we adopt a historical time step consisting of 40 steps with an interval of 10 s per step as input parameters. For ensuring safe shutdown, power reduction occurs at a rate of 5%FP/min starting from 100%FP, necessitating a shutdown duration of approximately 20 minutes. Consequently, an advance prediction step comprising 128 steps (equivalent to 1280 s) is determined to adequately fulfill the time advancement requirement for NPP fault warning.

The assumption is made that the data in the training and validation sets are independently and identically distributed, which is a commonly followed assumption in most feature selection studies. Firstly, the training set data were utilized for feature selection, with the optimal features being employed as input for the EFEM-BiLSTM algorithm. Subsequently, after each round of training on the training set, the prediction results are fed into the expert fuzzy evaluation module to assess the obtained prediction trend. The evaluation results are then provided as feedback to initiate the subsequent rounds of iterative training in the EFEM-BiLSTM model. In contrast, during validation on the validation set, no utilization of expert fuzzy evaluation module occurs.

The loss function value decreases as the number of training rounds increases during the prediction model training process. Generally, a smaller loss function indicates a more accurate model. However, deep learning models may encounter overfitting issues due to limitations in dataset size and network model complexity. Overfitting occurs when the trained model closely matches the training data but performs poorly on verification data, resulting in abnormal increases in the loss function value. To address this issue, random neuron inactivation is implemented in this study with a Dropout ratio set at 0.8.

In order to ensure the reliability of the prediction results, 100 repetitions are performed for each parameter trend during forecasting in the validation set. The final result is obtained by calculating the mean $\mu_Y$ of these 100 forecasting results. The upper bound of the 95% confidence interval is determined by adding $1.96\sigma_Y$ to $\mu_Y$, while the lower bound is determined by subtracting $1.96\sigma_Y$ from $\mu_Y$.

After undergoing 1000 rounds of training, the proposed prediction model yielded the following results for the 24 parameters：

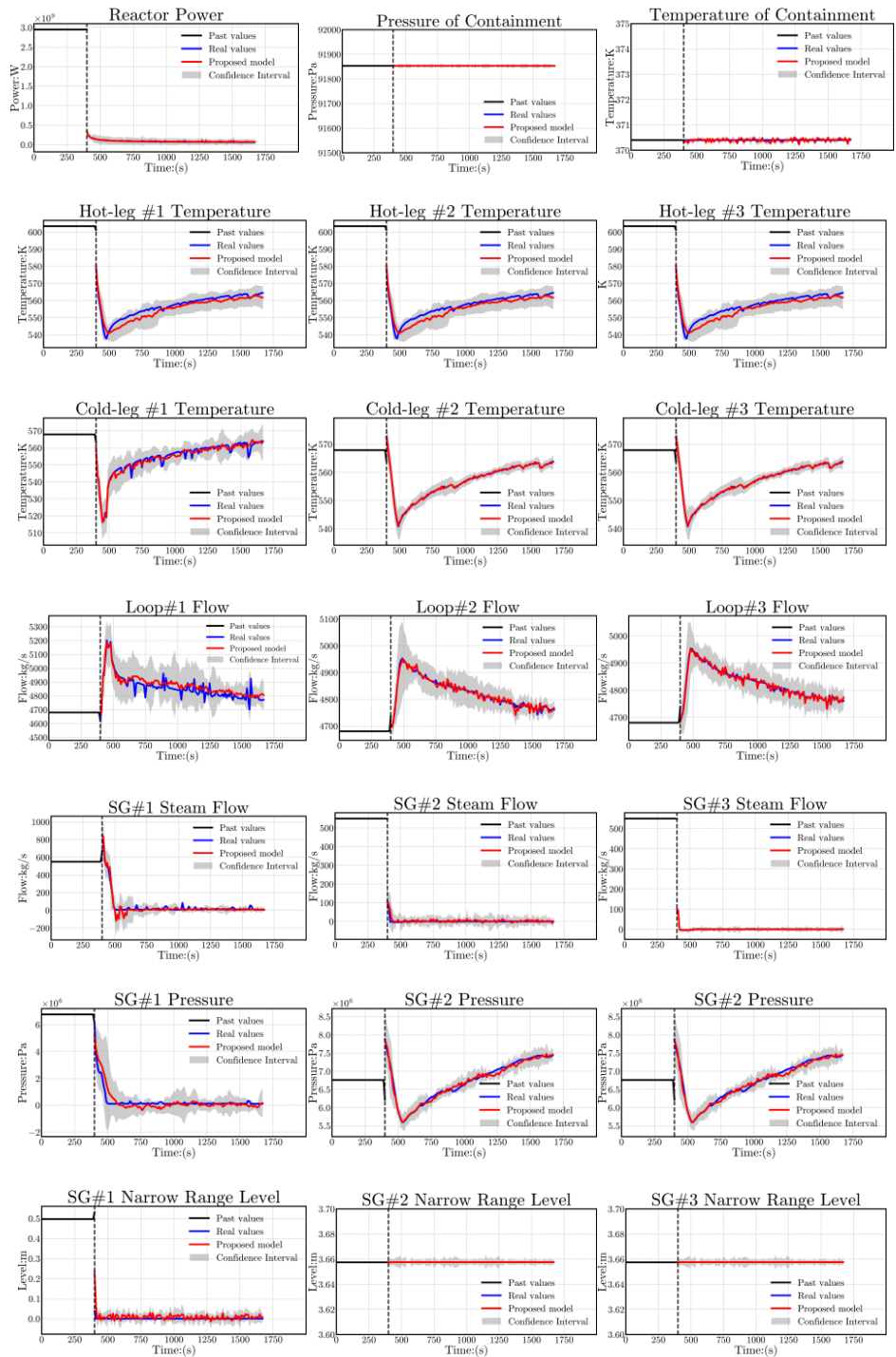

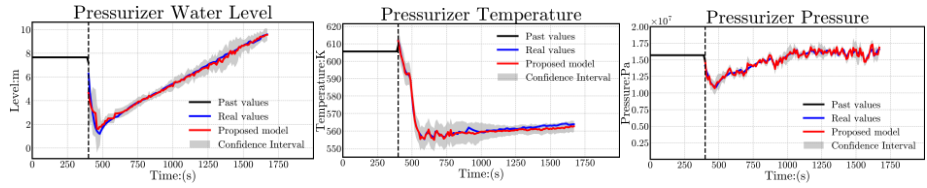

**Fig. 9**. Results for multiple parameters in 1280 s prediction test

with uncertainty estimation for 0.13 $m^2$ MSLB

The proposed prediction method is utilized to illustrate the curves of historical data, real data, and predicted data for 24 parameters in Fig. 9. It can be observed from Fig. 9 that, for the majority of parameters, the predicted data demonstrates a consistent trend with the real data, indicating a close proximity between the predicted values and actual values.

The EFEM-BiLSTM model effectively tracks and predicts the dynamic trends of crucial parameters during MSLB accidents in NPPs, even in the presence of significant variations such as Loop#3 Flow. Although there may exist certain deviations between the predicted data and actual measurements, these discrepancies remain within the acceptable range defined by the confidence interval.

4.5.2 Comparison

In order to further demonstrate the superiority of the adopted method, this study employs a range of evaluation indicators to comprehensively compare the performance of the other six models. The data is inputted into the corresponding control model after undergoing feature screening and normalization. To minimize potential confounding effects arising from disparate neuron counts and hyperparameters among different models, all hyperparameter values are set uniformly in this study. The parameters for each model are presented in Table 8.

Table 8 Model parameters

| Networks | | Input steps | Hidden units | Hidden layer num | Linear layer dimensions | Linear layer num | Output steps |
|---|---|---|---|---|---|---|---|
| Model 1 | RNN | 40 | 256 | 2 | - | - | 128 |
| Model 2 | LSTM | 40 | 256 | 2 | - | - | 128 |
| Model 3 | BiLSTM | 40 | 256 | 2 | - | - | 128 |
| Model 4 | Res-RNN | 40 | 256 | 2 | 520 | 2 | 128 |
| Model 5 | Res-LSTM | 40 | 256 | 2 | 520 | 2 | 128 |
| Model 6 | Res-BiLSTM | 40 | 256 | 2 | 520 | 2 | 128 |
| Model 7 | EFEM-BiLSTM | 40 | 256 | 2 | 520 | 2 | 128 |

Taking the cold-leg Temperature 1 (Cold-leg#1 Temperature) as an example, Fig 10 illustrates the prediction results obtained from each model. While Model 1 incorporates its own feedback mechanism to process historical information, it lacks the capability to effectively handle long-term data due to its indiscriminate selection of all historical information without memory screening. Consequently, there is a significant deviation between the predicted and actual data generated by Model 1. On the other hand, Model 2 represents an improved version of Model 1. When processing historical data, this approach utilizes selective forgetting to eliminate irrelevant information for precise predictions while effectively mitigating long-term dependency concerns. However, its limitation lies in its inability to incorporate bidirectional information fully and potential loss of specific characteristics. On the other hand, the BiLSTM model exhibits exceptional proficiency in capturing semantic dependencies from both forward and backward directions. Nevertheless, when predicting separately using models 1, 2, and 3, there might be initial deviations that require correction. In Fig.10, the initial deviation of the three models is reduced after incorporating a residual module to correct them. However, an issue still persists regarding error accumulation and gradual increase in deviation between predicted and actual data. The proposed model effectively addresses these challenges and exhibits significant advantages over other comparison methods.

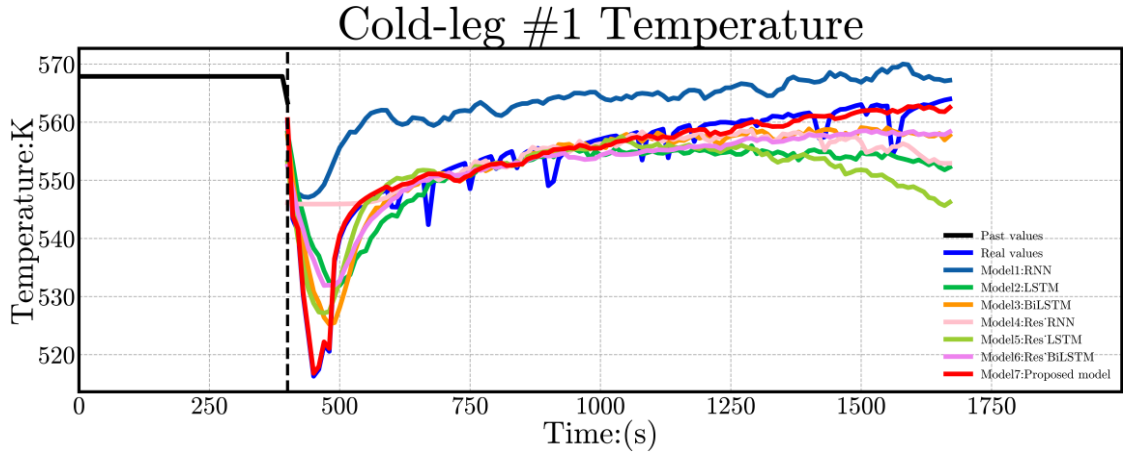

**Fig. 10.** Comparison of prediction results of Cold-leg#1 Temperature for 0.13 $m^2$ MSLB

The evaluation results of each model index on the same computing platform and data are presented in Table 9, aiming to visually demonstrate the disparity between the proposed method and the traditional approach.

Table 9 Results of model evaluation metrics

| Network | | Evaluation indicators | | | | | |
| --- | --- | --- | --- | --- | --- | --- | --- |
| | | MSE | RMSE | MAE | MAPE | DTW | TDI |
| Model 1 | RNN | 0.04980 | 0.2231 | 0.0872 | 24.582 | 7.651 | 17.412 |
| Model 2 | LSTM | 0.01256 | 0.1121 | 0.0358 | 12.621 | 5.793 | 17.582 |
| Model 3 | BiLSTM | 4.4365E-3 | 0.0667 | 0.0134 | 8.2721 | 5.179 | 19.631 |
| Model 4 | Res-RNN | 3.5729E-3 | 0.0597 | 0.0276 | 14.561 | 2.294 | 15.857 |
| Model 5 | Res-LSTM | 2.3104E-3 | 0.0480 | 0.0178 | 6.0325 | 2.235 | 15.462 |
| Model 6 | Res-BiLSTM | 2.2327E-3 | 0.0482 | 0.0114 | 5.6721 | 1.577 | 16.584 |
| Model 7 | EFEM-BiLSTM | 1.3069E-3 | 0.0361 | 0.0036 | 1.6667 | 0.257 | 2.995 |

The MSE index calculates the sum of squared differences between corresponding elements of the predicted sequence and the true sequence, thereby quantifying the spatial disparity between the two sequences. A smaller value signifies a higher degree of concordance between them. The RMSE index is derived as the square root of the MSE index. MAE represents the average modulus length of errors, disregarding their direction, by summing up absolute differences between predicted values and true values. MAPE quantifies time series prediction accuracy, with a perfect prediction indicated by a 0% value and an accurate prediction suggested by a MAPE value below 10%. The DTW index measures local similarity between two sequences through localized warping and matching along the time axis. TDI serves as an evaluation metric within the DILATE loss function proposed by Le Guen and Thome (2019), assessing temporal inconsistencies such as lag predictions between predicted and ground truth sequences.

The proposed model surpasses all the compared models, as demonstrated in Table 7. It exhibits the lowest overall matching degree among all models across four indicators: local matching degree evaluated by DTW, time deviation degree of TDI, RMSE, MSE, MAE and MAPE. However, solely relying on the residual mechanism fails to significantly enhance the model's performance and overcome local minima during training. Using only the residual mechanism in LSTM can merely ensure that its performance is not inferior to pure LSTM but cannot guide model training or facilitate escape from local minima.

The findings demonstrate that the proposed model accurately predicts parameter changes up to 128 steps in advance, equivalent to a duration of 1280 s. These predictive outcomes serve as reliable references for operators when making informed decisions.

## 5 Summary and Discussion

Existing NPPs designs lack inherent safety features, and the possibility of serious accidents cannot be completely eliminated. Early detection of potential faults and timely scheduling of maintenance activities can significantly mitigate operational risks of power plants and enhance the reliability of operators' decision-making during emergency response processes. This paper addresses the multi-step prediction problem of PHM in NPPs using a reinforcement learning approach. By integrating expert fuzzy evaluation with Long Short-Term Memory (LSTM) neural networks, the proposed model predicts future parameter trends based on their historical information. The model undergoes self-optimization during training to minimize cumulative errors generated during the prediction process.

In this study, the CPR1000 NPP simulator was employed to simulate and collect data under MSLB accident conditions with 20 different rupture sizes. These simulated data were subsequently utilized for training and validating the proposed method. Initially, feature selection was conducted to eliminate redundant features and obtain optimal ones. The optimal features were then normalized before being inputted into various comparative models. The performance of six different models on the same dataset and computing platform was compared, demonstrating that the proposed model can provide fault warning and prediction based on plant monitoring parameters, thereby mitigating the risk of MSLB accidents that pose threats to system safety or cause severe shutdowns.

The proposed prediction method can achieve early prediction of multiple parameters 128 steps ahead (i.e., 1280 s) under accident scenarios, thereby providing sufficient information for reliable decision-making by NPP operators and improving the overall safety of system operation. Therefore, this prediction model is suitable for practical application in NPPs. Moreover, this method also offers an effective reference solution for PHM applications such as anomaly detection and remaining useful life prediction.

The proposed prediction method has currently only been validated for the Main Steam Line Break (MSLB) accidents. In future work, we will conduct testing and validation of the proposed model for additional fault scenarios. Furthermore, we aim to enhance the algorithm by incorporating the physical coupling relationships between parameters to achieve earlier predictions.

## Declaration of Competing Interest

The authors declare that they have no known competing financial interests or personal relationships that could have appeared to influence the work reported in this paper.

## Acknowledgment

This work was supported by the Science and Technology Project of Fujian Province (No. 2022H0004).